\documentclass{article}

\usepackage{PRIMEarxiv}

\usepackage[utf8]{inputenc} 
\usepackage[T1]{fontenc}    
\usepackage{hyperref}       
\usepackage{url}            
\usepackage{booktabs}       
\usepackage{array}
\usepackage{amsfonts}       
\usepackage{nicefrac}       
\usepackage{microtype}      
\usepackage{lipsum}
\usepackage{fancyhdr}       
\usepackage{graphicx}       
\graphicspath{{media/}}     
\usepackage{amsmath}
\usepackage{tcolorbox}
\usepackage{array}
\usepackage{graphicx}

\title{FinTeamExperts: Role-Specialized MOEs for Financial Analysis}

\author{
 Yue Yu \\
 Ney York University \\
 \texttt{yy1879@nyu.edu} \\
 \And 
 Prayag Tiwari\\
 Halmstad University, Sweden \\
 \texttt{prayag.tiwari@hh.se} \\
}

\begin{document}
\maketitle








\begin{abstract}

Large Language Models (LLMs), such as ChatGPT, Phi3 and Llama-3, are leading a significant leap in AI, as they can generalize knowledge from their training to new tasks without fine-tuning. However, their application in the financial domain remains relatively limited. The financial field is inherently complex, requiring a deep understanding across various perspectives, from macro, micro economic trend to quantitative analysis. 

Motivated by this complexity, a mixture of expert LLMs tailored to specific financial domains could offer a more comprehensive understanding for intricate financial tasks. In this paper, we present the FinTeamExperts, a role-specialized LLM framework structured as a Mixture of Experts (MOEs) for financial analysis. The framework simulates a collaborative team setting by training each model to specialize in distinct roles: Macro Analysts, Micro Analysts, and Quantitative Analysts. This role-specific specialization enhances the model's ability to integrate their domain-specific expertise. We achieve this by training three 8-billion parameter models on different corpus, each dedicated to excelling in specific finance-related roles. We then instruct-tune FinTeamExperts on downstream tasks to align with practical financial tasks. The experimental results show that FinTeamExperts outperform all models of the same size and larger on two out of four datasets. On the stock prediction, which presents a more complex task, FinTeamExperts still surpass all models of the same size. This highlights the success of our role-based specialization approach and the continued training approach for FinTeamExperts.

\end{abstract}



\keywords{LLMs \and Financial Analysis \and Mixture of Experts }

\section{Introduction}

\begin{figure}
    \centering
\includegraphics[width=0.9\textwidth]{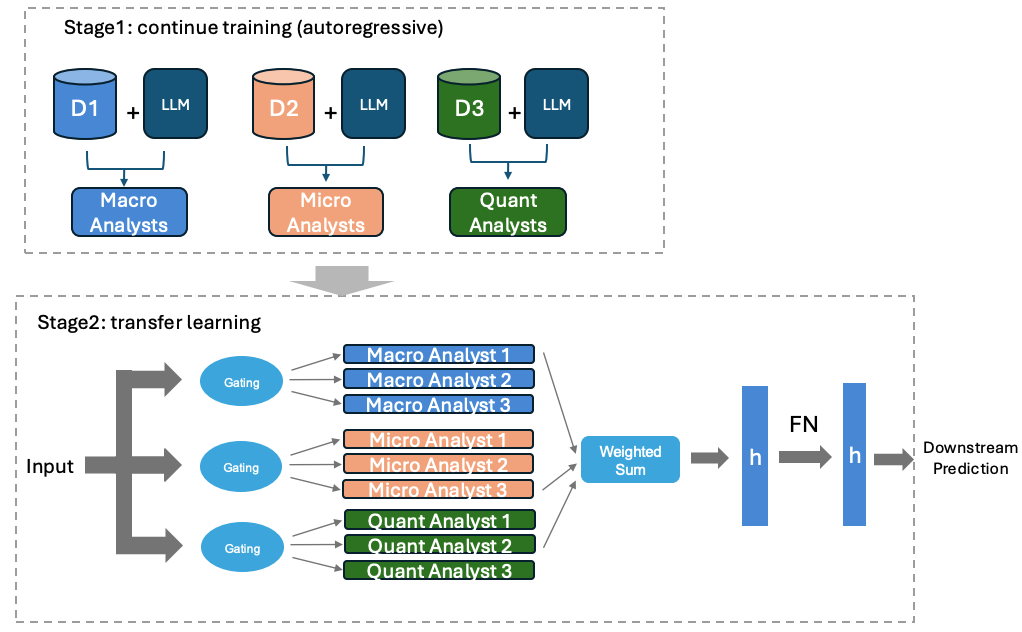}
    \caption{Architecture of FinTeamExperts with role specialized MOEs for financial analysis.}
    \label{fig:architect}
\end{figure}

Traditional machine learning methods, such as Support Vector Machines (SVMs) \cite{xue2009svm}, gradient-boosted trees \cite{ye2009stochastic}, and logistic regression \cite{lavalley2008logistic}, have limitations when it comes to understanding, reasoning, and generalizing in language-specific tasks. This limitation was overcome with the introduction of transformer architecture \cite{subakan2021attention}, which relies solely on attention mechanisms, removing the need for recurrence and convolutions. By training on large datasets using autoregressive techniques, Large Language Models are able to capture context and semantic dependencies in language. They are highly effective in translating across languages, processing large volumes of data, delivering quick responses with minimal delay, and can be fine-tuned to handle specific tasks and domains.

LLMs are increasingly used in the finance industry. They improve customer service with chatbots, help summarize information, recommend relevant knowledge, and automate tasks like filling out forms. LLMs also support risk management by analyzing market and credit risks and detecting anomalies or sentiment analysis. In investment, they assist with quantitative analysis, and in document processing, they help ensure compliance with regulations. However, research in financial Large Language Models (FinLLMs) remains limited. The earlier model, FinBERT\cite{liu2021finbert}, adapts BERT\cite{devlin2018bert} for financial sentiment analysis by pre-training on financial documents and fine-tuning with a sentiment-specific dataset, achieving superior results but struggling with tasks beyond sentiment analysis. FinMA\cite{xie2023pixiu} fine-tuned Llama with financial instructions, showing competitive performance but not surpassing larger models like GPT-4. 

Despite these attempts in the financial domain, the intricate and domain-specific nature of financial tasks poses challenges that general-purpose LLMs are not well-suited to address. 
To bridge this gap, we introduce FinTeamExperts, a novel framework of LLM designed as a Mixture of Experts (MoEs) \cite{cai2024survey} to excel in finance-specific tasks. FinTeamExperts consist of three 8-billion-parameter models, each carefully tailored to comprehend and generate content relevant to the finance domain. MoEs, techniques that are used to enhance LLM performance without increasing computational demands, is used in the proposed framework to improve the domain-specific performance. This framework draws inspiration from real-world team dynamics, where each member develops expertise in distinct areas. By employing this specialized approach, we aim to create a powerful model capable of effectively handling the complex and nuanced demands of financial analysis and decision-making.

Our methodology follows a two-phase training process, as shonw in the Figure \ref{fig:architect}. In the first phase, the models are pre-trained on a curated corpus of role-related data to establish a strong foundational understanding of financial concepts. In the second phase, models are combined using routing gates and undergo instruct-tuning. This fine-tuning process helps the models produce outputs that are well-aligned with real-world tasks in the finance domain. The supervised instruct-tuning ensures that FinTeamExperts are skilled at applying knowledge from different experts to different real-world scenarios, making them effective for complex financial analysis and decision-making tasks.

We evaluate FinTeamExperts on four public datasets, where it consistently outperforms models of the same size. These results demonstrate the model’s superior performance across a range of finance-specific tasks, highlighting the effectiveness of the proposed team experts settting and the specialized two phase training framework.

FinTeamExperts represent a significant advancement in the MOE framework of large language models to the financial domain. By offering domain-specific roles and training methodology, the framework provides a valuable tool for financial professionals, enhancing the performance and reliability of financial analysis and decision making in complex real-world scenarios.

We summarize our contributions as follows:

\begin{itemize}
    \item A novel real-world role-based framework, FinTeamExperts, where each expert is specialized in a distinct financial role.
    \item An adapted MoE framework that enables team-based specialization, enhancing task-specific performance.
    \item The framework’s effectiveness is validated through extensive experiments, showcasing superior performance on finance-related tasks.
    \item An ablation study that demonstrates the individual contributions of each expert model within the framework.
\end{itemize}

\section{Methodologies}
In this section, we present the framework, outline the roles within the team, explore the adaptation of learning processes, and discuss the architectures of MOEs. 

The FinTeamExperts architecture consists of two stages designed for financial analysis using a MoE approach. In stage one of continue autoregressive training, LLMs are fine-tuned with different datasets (marked as D1, D2, D3 in Figure \ref{fig:architect}) to create specialized models, referred to as Experts. Each expert model is trained on its respective dataset, allowing it to specialize in a specific domain or task relevant to financial analysis. This phase focuses on autoregressive training, where the experts continue with predicting the next token in the given corpus.

In stage two of transfer learning, the trained experts are applied together to new financial tasks. The input is passed through a gating mechanism, which selects the most appropriate experts for the task at hand. The outputs from these selected experts are combined via a weighted manner and processed by hidden layers and a feedforward network, which refine the aggregated information. This setup enables the model to generalize and make downstream predictions for financial applications, such as trend analysis or risk assessment.

\subsection{Framework Formulation}
The core of the MOE is the routing mechanism of inputs to Experts. We set up a group of experts, corresponding to macro experts, micro experts, and quant experts. A hard gating mechanism is used to select one expert from each group, and their outputs are then weighted via another soft gating mechanism. The routing gate can be expressed as $g(x_i)g(x_i) = \big(g_1(x_i), g_2(x_i),\dots,g_K(x_i)\big)$ representing a probability distribution over the experts.
\begin{equation}
g(x_i) = \text{Softmax}(W_g \cdot x_i)
\end{equation}
where $x_i$ is the input token and $W_g$ is the gating network weight matrix, which learns to route inputs to appropriate experts.

\begin{equation}
    y_i = \sum_{j=1}^{K} g_j(x_i) E_j(x_i)
\end{equation}

In a simplified three-role gating framework, we can denote the output $y_i$ for an input token $x_i$, as a weighted sum of the outputs from the selected experts. Each expert $E_j(x_i)$ of expert $j$ produces a unique output for teh input $x_i$. $K$ is the total number of experts available, for instance, three in our study. Typically, only a subset of these experts is used for each input but we have it as a weighted contribution.

To build a adaptable framework where a flexible number (e.g., K) of experts caon contribute to each expert category. The weighted output $y_i$ is thus adjusted as a sum over all roles and their respective experts,  with each category weighted by a corresponding gating functions. This is defined as:
\begin{equation}
   y_i = \sum_{r \in \{\text{Macro}, \text{Micro}, \text{Quant}\}} \sum_{j=1}^{K} g_{r}(x_i) \cdot h_{r,j}(x_i)\cdot E_{r,j}(x_i)
\end{equation}
where $r$ denotes the role (e.g., Macro, Micro, Quant), and $g_{r,j}(x_i)$ is the gating function that assigns weights to each expert. In particular, we set 
\begin{equation}
    h_{r,j} = \text{GumbelSoftmax}(W_g \cdot x_i),
\end{equation}
for all $r$ to select one expert from each group of experts. 

For inference, we define the inference head $\hat{y}_t = C(y_{j<t})$, which processes outputs from previous tokens to generate the next prediction.

The overall loss for the FinTeamMOE model, $\mathcal{L}_{\text{task}}$, includes both the task-specific loss and a regularization term. The task-specific loss, such as cross-entropy, is used for training the model on the primary task:
\begin{equation}
\mathcal{L}_{\text{MOE}} = \mathcal{L}_{\text{task}} + \lambda \cdot H(g(x))
\end{equation}
 
where $\lambda$ is a regularization coefficient controlling the influence of the entropy term $H(g(x))$. This entropy term promotes balanced utilization of experts by encouraging the gate function to distribute weights across experts effectively.0

\subsection{Teamed Roles}
To optimize performance and achieve strategic goals, FinTeamExperts focus on three pivotal roles within the investment setting:

\textbf{Macro Analysts}: Macro analysts study broader economic trends, geopolitical events, and global market movements to inform investment strategies. Their insights are crucial for understanding the larger economic context in which specific investments operate, enabling more informed decision-making.

\textbf{Micro Analyst}: Portfolio managers are responsible for making investment decisions and managing a portfolio of assets. They analyze market trends, assess risk, and determine the best investment strategies to maximize returns while adhering to the company’s risk tolerance and investment objectives. Their role is vital in driving the overall investment strategy and ensuring alignment with the company’s financial goals.

\textbf{Quantitative Analysts}: Quants develop mathematical models and algorithms to analyze financial data and identify trading opportunities. They work on creating predictive models, optimizing trading strategies, and automating trading processes using statistical and computational techniques. In the age of data-driven decision-making, quants provide the technical expertise needed to enhance trading efficiency, manage risks, and uncover alpha-generating opportunities.

Together, these roles form a comprehensive team within FinTeamExperts that balances strategic decision-making, technical innovation, and economic analysis, driving success in the competitive landscape of trading companies.

\subsection{Adapting LLM Knowledge}

We continue training the LLM checkpoints (such as GPT-2 and LLaMA-3-8B) using typical autoregressive learning with the next token prediction as the learning objective, defined as:

\begin{equation}
P(x_1, x_2, \dots, x_T) = \prod_{t=1}^{T} P(x_t \mid x_1, x_2, \dots, x_{t-1})
\end{equation}

where $x_1, x_2, \dots, x_T$ are the tokens from the corpus. The model generates each token $x_t$ conditioned on the previous tokens.

Adapting large language models to specialize in finance-domain tasks involves curating a comprehensive financial corpus and implementing role-specific training. The initial phase includes pretraining the model on a curated dataset of market reports, economic analyses, and financial statements, building a robust understanding of financial terminology and concepts. Each model within FinTeamExperts is then trained to specialize in one of the three roles, developing expertise in macroeconomic analysis, asset management, or statistical trading techniques. This specialized training enables the models to integrate their expertise, forming a comprehensive financial analysis tool.

\subsection{Mixture-of-Experts Architecture}
The FinTeamExperts framework leverages the Mixture of Experts (MOEs) architecture to optimize performance in finance-related tasks. MOEs utilize multiple expert models, each specializing in different aspects of financial analysis, and dynamically route queries to the most relevant expert. This architecture allows for efficient resource allocation and enhances the model's ability to handle diverse and complex financial scenarios.

We employ Dynamic Routing: Queries are dynamically routed to the most appropriate expert model based on the specific financial task, improving accuracy and efficiency.

Specialized Training: Each expert model undergoes specialized training focused on its designated role (Macro Analysts, Portfolio Managers, Quantitative Analysts), ensuring depth of knowledge and proficiency in specific areas.

Hierarchical Expertise: The architecture supports hierarchical expertise, where higher-level experts oversee and refine the outputs of lower-level models, ensuring coherent and high-quality analysis.
This MOEs-based methodology allows FinTeamExperts to leverage the strengths of individual expert models while maintaining flexibility and adaptability in addressing a wide range of financial tasks. The integration of these innovations demonstrates the potential of advanced LLMs in transforming financial analysis and decision-making.

\section{Experiments}
\subsection{Role-specific Continue Training Datasets}
\label{ss:roles}
In this study, we utilize three categories of datasets to continue pretrain models for financial roles: Macro of financial News, Financial Statements and Reports, and Market and Transactional Data. Each category serves a distinct role in equipping the model with the necessary knowledge to support the varied analytical needs of macroeconomic, microeconomic, and quantitative investment strategies.

\textbf{Financial News Datasets:} This category includes datasets that capture the sentiment and content of financial news articles, providing valuable insights into market trends and investor sentiment. One notable resource is the Thomson Reuters News Analytics (TRNA) dataset, which offers sentiment data derived from a vast archive of news articles. These datasets are crucial for developing models that can interpret the influence of global events and market news on investment decisions.

\textbf{Financial Statements and Reports Datasets:} This category encompasses structured financial data extracted from corporate reports such as balance sheets, income statements, and cash flow statements. The SEC EDGAR database \footnote{https://www.sec.gov/} provides access to detailed financial disclosures filed by publicly traded companies in the U.S., including annual (10-K) and quarterly (10-Q) reports. Similarly, the WRDS database \footnote{https://wrds-www.wharton.upenn.edu/} offer comprehensive financial statement data from global companies. These datasets enable models to perform detailed microeconomic analysis, assessing the financial health and performance of individual companies.

\textbf{Market and Transactional Data Datasets:} This category includes datasets containing historical and real-time market data, such as stock prices, trading volumes, and other transactional data. The NASDAQ Data \footnote{https://data.nasdaq.com/} provides a wide range of financial, economic, and alternative data, while the Alpha Vantage API offers access to real-time and historical market data across various asset classes. Additionally, Yahoo Finance \footnote{https://www.yahoofinanceapi.com/} offers extensive historical market data and financials, which are essential for developing quantitative models that optimize trading strategies and identify market opportunities.

By integrating these diverse datasets, we create a robust foundation for training models that can navigate the complexities of financial markets, supporting the analytical needs of macro analysts, micro analysts, and quantitative analysts alike.

\subsection{Downstream Datasets}
FPB dataset \cite{Malo2014GoodDO} contains 4,840 sentences from English-language financial news articles, each labeled by sentiment. Sentences are categorized based on the level of agreement among 5 to 8 annotators, providing a clear indication of consensus in sentiment labeling.

\begin{table}[h]
\centering
\begin{tabular}{p{2.5cm} p{12cm}}
\toprule
\textbf{Speaker} & \textbf{Dialog} \\
\midrule
User & What is the sentiment of the following financial post: Positive, Negative, or Neutral? \newline Text: What's up with \$LULU? Numbers looked good, not great, but good. I think conference call will instill confidence.  \\
\midrule
Assistant & The sentiment is neutral. \\ 
\midrule
User & Examine the data and tweets to deduce if the closing price of \$cvx will boost or lower at 2017-02-21. Kindly confirm either Rise or Fall. \newline 
Context: date,open,high,low,close,adj-close,inc-5,inc-10,inc-15,inc-20,inc-25,inc-30 \newline 
2017-02-06,0.3,0.5,-0.5,-0.5,-0.5,-0.7,0.6,1.2,1.6,2.1,2.5 \newline 
2017-02-07,1.4,1.5,-0.3,-1.4,-1.4,0.8,1.6,2.4,2.8,3.3,3.7 \newline 
2017-02-08,-0.4,0.2,-1.1,0.2,0.2,0.7,0.9,2.0,2.5,2.9,3.3 \newline 
\ldots \newline
2017-02-06: \#dividend growth investing at \#work - the streak continues! \#investing \#dividends \$cvx \newline 
rt AT\_USER chevron corporation \$cvx shares bought by los angeles capital management \& equity research inc. \newline
2017-02-07: rt AT\_USER 4 energy stocks that are ticking time bombs \$cvx \$oxy \$cop \newline 
rt AT\_USER mike liss of AT\_USER century (\$twvlx) talked about \$cvx \$apc \$mdlz \$mdt \& \$mrk during "hold it or fold it" \newline 
2017-02-08: rt AT\_USER dynamic capital management ltd reduces stake in chevron corporation \$cvx \newline 
dynamic capital management ltd reduces stake in chevron corporation \$cvx \newline 
rt AT\_USER \#chevron partners zinox for co \newline 
2017-02-09: Answer: \\

\midrule 
Assistant & The prediction is Rise. \\
\bottomrule
\end{tabular}
\caption{Example of a dialog between the user and the assistant.}
\label{tab:dialog_example}
\end{table}

The FLARE-FIQASA dataset \cite{flare_fiqasa_2023} is a labeled collection of financial texts, such as social media posts and headlines, used for sentiment analysis. It categorizes each text as positive, negative, or neutral to help analyze market sentiment in finance-focused content.

FinQA is designed for numerical reasoning over financial data, integrating both textual and tabular information. It contains examples with pre-text, post-text, and tables, along with associated questions and reasoning programs to compute numerical answers.

FOMC dataset \cite{shahetal2023trillion} is a hawkish-dovish classification task involves categorizing statements based on sentiment toward monetary policy. 'Hawkish' statements signal a preference for tightening policy to control inflation, while 'dovish' ones indicate support for accommodative measures to boost growth. This classification is challenging due to the nuanced language in FOMC statements, which influences financial markets and economic expectations.

\begin{table}[ht]
\centering
\begin{tabular}{l c c c c}
\toprule
\textbf{Model} & \textbf{Size} & \textbf{FPB} & \textbf{FiQA-SA} & \textbf{FOMC} \\
\midrule
FinMA-Full \cite{xie2023pixiu} & 7B & 87.0 & 79.0 & - \\
GPT4\cite{achiam2023gpt} & - & 78.0 & \underline{80.0} & \textbf{71.0} \\
BLOOM \cite{le2023bloom} & 176B & 50.0 & 53.0 & - \\
BloombergGPT \cite{wu2023bloomberggpt} & 50B & 51.0 & 75.0 & - \\
Llama-3-8B \cite{dubey2024llama} & 8B & 84.95 & 65.1 & 59.2 \\
Qwen-2-7B \cite{yang2024qwen2} & 7B & 52.0 & 57.0 & 63.0\\
Mistral-8$\times$7B-v1 \cite{jiang2024mixtralexperts}& 7B & 29.0 & 16.0 & 37.0 \\
\midrule
FinTeamExpert & 3$\times$1B & \underline{89.3} & 76.3 & 64.2 \\
FinTeamExpert & 3$\times$8B & \textbf{90.5} & \textbf{81.0} & \underline{66.5} \\
\bottomrule
\end{tabular}
\caption{Main Results of FinTeamExperts for sentiment tasks}
\label{table:main_results}
\end{table}






\begin{table}[ht]
\centering
\begin{tabular}{lcc}
\toprule
\textbf{Model} & \textbf{Size} & \textbf{CIKM18} \\
\midrule
ChatGPT & - & 55.0 \\
GPT4 \cite{achiam2023gpt} & - & \textbf{57.0} \\
Gemini \cite{team2023gemini} & - & 54.0 \\
Qwen2-7B \cite{yang2024qwen2} & 7B & 52.0 \\
FinMA 7B-full \cite{xie2023pixiu} & 7B & 53.0 \\
Llama-3-8B \cite{dubey2024llama} & 8B & 51.9 \\
Mistral-8$\times$7B-v1 \cite{jiang2024mixtralexperts} & 7B & 42.0 \\
\midrule
FinTeamExpert & 3$\times$1B & 50.3 \\
FinTeamExpert & 3$\times$8B & \underline{56.3} \\
\bottomrule
\end{tabular}
\caption{\label{tb:res_cikm18}Results on stock prediction dataset}
\label{table:ablation_study}
\end{table}

\subsection{Settings}
We build our expert model, on top of GPT-3-Large (with less than 1B parameters), noted as \texttt{E-1B}, and LLaMA-3-8B, noted as E-8B. 

\texttt{E-1B} consists of 24 layers, each with 16 attention heads and a hidden size of 1,536, while \texttt{E-8B} comprises 36 layers, 32 attention heads, and a hidden size of 4,096. Using pretrained weights as the backbone for both models, we further train them on their respective corpora as specified in section \ref{ss:roles}.

For \texttt{E-1B}, we use a learning rate of $1 \times 10^{-4}$ with 500 warmup steps, a cosine learning rate scheduler, and the Adam optimizer \cite{kingma2017adammethodstochasticoptimization}, with $\beta_1 = 0.9$, $\beta_2 = 0.95$, and a weight decay of 0.1. The model is trained for one epoch.

For \texttt{E-8B}, we use a learning rate of $1 \times 10^{-4}$ with 1,000 warmup steps, a cosine scheduler, and the Adam optimizer with $\beta_1 = 0.9$, $\beta_2 = 0.95$, and an epsilon value of $1 \times 10^{-6}$. We apply a weight decay of 0.1 and train the model for one epoch.

Both models are trained using 16-bit mixed precision on four 24GB A10G GPUs with fully sharded data parallelism, implemented via the Accelerate library\footnote{\url{https://huggingface.co/docs/accelerate}}.


\subsection{Reseults}
The FinTeamExpert models, particularly the 3×8B version, achieve the best performance across all sentiment tasks, with the highest scores in FPB (90.5), FiQA-SA (81.0), and FOMC (65.5). This suggests that increasing model size and using an ensemble of models enhances performance in financial sentiment analysis. Even the smaller FinTeamExpert (3×1B) performs competitively, especially in FPB (89.3) and FOMC (64.2), demonstrating that well-structured ensembles of smaller models can rival larger models in specialized tasks.

In comparison, FinMA-Full (7B) performs well on FPB (87.0) and FiQA-SA (79.0), though it falls slightly behind the FinTeamExpert models. On the other hand, GPT-4, while a powerful general-purpose model, lags behind in financial tasks with a lower FPB score (78.0). Large models like BLOOM (176B) and BloombergGPT (50B) underperform in FPB, highlighting that model size alone does not guarantee effectiveness without domain-specific adaptation.

LLaMA-3-8B (8B) shows strong results, particularly on FPB (84.95), demonstrating solid performance compared to similarly sized models. Overall, FinTeamExperts are the most effective across all tasks, indicating that specialized, ensemble-based models outperform general-purpose models in the financial sentiment domain.

Table \ref{tb:res_cikm18} shows the resullt on stock prediction dataset. GPT-4 leads the CIKM18 task with a score of 57.0, closely followed by the domain-specific FinTeamExpert (3×8B) at 56.3, showing that both general-purpose and specialized ensemble models can excel in financial tasks. FinMA (7B) and LLaMA-3-8B achieve scores of 53.0 and 51.9, respectively, demonstrating solid performance but falling short of the top two models. 

The smaller FinTeamExpert (3×1B) scores 50.3, highlighting that even smaller ensemble models remain competitive, though not as strong as their larger counterparts. Overall, GPT-4 and FinTeamExpert (3×8B) dominate, while other models offer reasonable results in the CIKM18 task.

\subsection{Ablation Study}

We conduct a perplexity analysis on three role-oriented training models and our FinTeamExperts model, using mini-test samples to assess their perplexity scores. Figure \ref{fig
} illustrates these results, with a benchmark perplexity score of approximately 15 for the vanilla Llama-3-8B model before it was trained with role-specific adaptations. Perplexity trends are plotted for the macro, micro, and quant models, while downstream learning results are shown for FinTeamExperts. As training progressed, perplexity scores for all three role-oriented models decreased and eventually converged around 6.5. Among these, the macro-role model achieved the lowest perplexity, with the quant-role model slightly higher, though the differences were minor.

FinTeamExperts, routed and fine-tuned for downstream financial tasks, began with an initial perplexity below 8 and converged to a value under 6. This indicates that FinTeamExperts is better adapted to the financial dataset, demonstrating enhanced understanding and alignment with the financial tasks.

\begin{figure}
    \centering
    \includegraphics[width=0.7\linewidth]{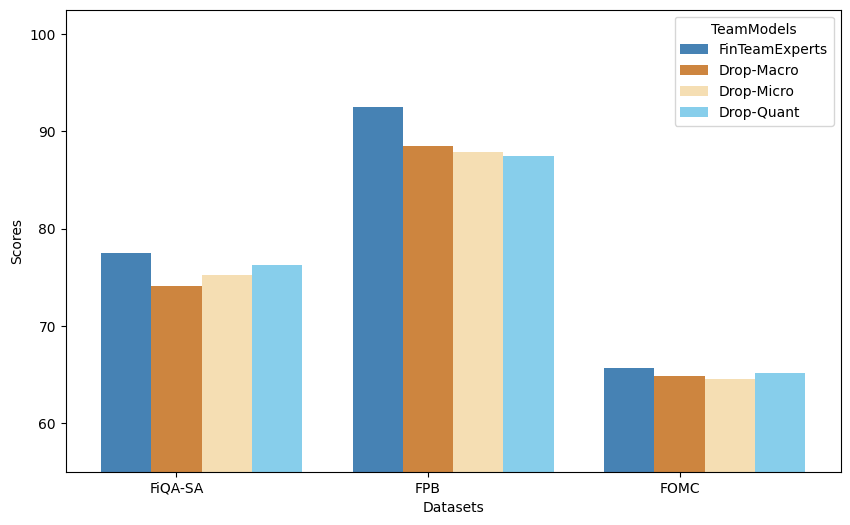}
    \caption{Ablation study of drop-one analysis}
    \label{fig:drop-one}
\end{figure}

The ablation study, as shown in Figure \ref{fig:drop-one}, reveals how removing individual roles—Macro, Micro, or Quant—impacts model performance across different tasks. The results clearly show that the full FinTeamExperts, without any role removed, consistently outperforms all ablated versions, particularly in complex tasks like FPB. This highlights the critical importance of each role in achieving top performance. For InFiQA-SA, dropping any role yields similar performance, suggesting that no single role is indispensable for this task. However, in FPB, the complete model far exceeds any version with a role removed, underscoring the collaboration required for optimal results. For FOMC, droping Micro lead to lower performance, indicating the importance of micro knowledge of this task. Overall, these findings reinforce the value of having all roles intact for the most challenging tasks.
\begin{figure}
    \centering
    \includegraphics[width=0.7\linewidth]{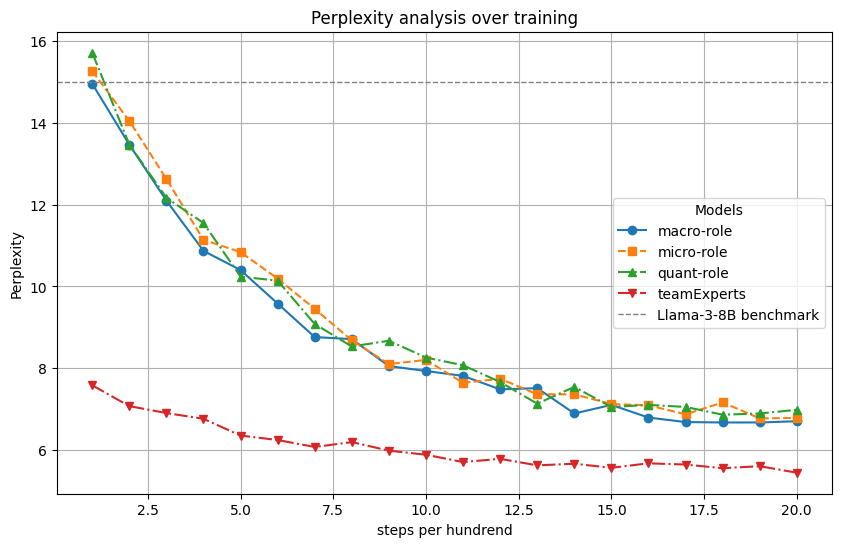}
    \caption{Perplexity analysis}
    \label{fig:perplexity}
\end{figure}

\section{Related Works}
\subsection{Financial LLMs}

Prior to the rise of LLMs, deep learning played a pivotal role in the financial domain, particularly in tasks like portfolio management \cite{zhang2020deep} and risk management \cite{mashrur2020machine}. In portfolio management, deep learning can be utilized to enhance the optimization process by directly maximizing metrics such as the Sharpe ratio. This approach simplifies asset selection by effectively capturing correlations across different asset classes, thereby streamlining decision-making in constructing a balanced portfolio.

In the era of advanced LLMs, research specifically targeting financial language models (FinLLMs) remains relatively limited, as highlighted by a recent survey \cite{lee2024survey}. One notable example is FinBERT \cite{liu2021finbert}, which adapts the general-purpose BERT model for financial sentiment analysis using a two-step approach. First, it is pre-trained on financial texts, utilizing a subset of the Reuters TRC2 dataset to enhance its understanding of financial terminology. Next, a dense layer is added to the final hidden state of the classification token (CLS), and the model is fine-tuned using the Financial PhraseBank (FPB) dataset. FinBERT achieves outstanding performance in financial sentiment analysis, surpassing state-of-the-art models in this specific task. However, it remains limited in scope and does not perform as effectively on other financial tasks.

FinGPT \cite{yang2023fingpt} provides an end-to-end framework for training and applying financial large language models in the finance industry. It uses the efficient Low-rank Adaptation (LoRA) \cite{hu2021lora} technique to fine-tune open-source models like LLaMA and ChatGLM with around 50,000 samples. However, its evaluation is currently limited to finance classification tasks only.

BloombergGPT \cite{wu2023bloomberggpt}, a 50-billion parameter model based on BLOOM’s architecture, is one of the first decoder-only LLMs trained specifically for finance. Trained from scratch on 363 billion tokens from financial documents and 345 billion from general datasets, it predicts the next token in documents without fine-tuning on instructions. However, its results lag behind those of other models, including some much smaller ones, as detailed in \cite{rodriguez2024large}.

FinMA \cite{xie2023pixiu} curated an instruction dataset for a financial LLM and fine-tuned it on Llama, resulting in strong performance among similar-sized LLMs, though not surpassing larger models like GPT-4.

\subsection{Mixture of Experts}
Mixture of Experts (MoEs), derived from Gaussian Mixture models, are employed to boost the performance of large language models (LLMs) without increasing computational resource requirements. 

SwithTransformer \cite{fedus2022switch},  places MoE layers after the multi-head attention mechanism in each transformer block to select the feedforward layers. Unlike LLMs where all parameters are used for every input, SwitchTransformer activates only a small subset of experts for each input, significantly reducing computational costs. A routing network determines which expert should handle each input, ensuring that only the most relevant parts of the model are utilized for each task.

GLaM \cite{du2022glam}, a model with 1.2 trillion parameters, is approximately 7 times larger than GPT-3, yet it achieves this scale with only a third of the energy consumption required to train GPT-3. GLaM's architecture integrates MoE layers with Transformer layers. For each input token, a gating module selects the two most relevant experts, and a weighted average of their outputs is passed to the next Transformer layer. The gating network, which is trained to identify the optimal experts for each token in the input sequence, ensures efficient use of computational resources while maintaining high performance.

ST-MoE \cite{zoph2022st}, is a sparse Transformer model with 269B parameters, offering performance at a computational cost similar to that of a dense 32B parameter encoder-decoder Transformer. The model recommends a top-2 routing mechanism, with each input token routed to the two most relevant experts, ensuring computational efficiency. A capacity factor of 1.25 is recommended, which controls the number of tokens processed by each expert, and this factor can be adjusted during evaluation to meet changing memory or computational requirements. Additionally, quality improvements are achieved through dense layer stacking and the introduction of a multiplicative bias.

\section{Conclusion}
In this paper, we introduced FinTeamExperts, a novel framework of role-specialized large language models (LLMs) designed as a Mixture of Experts (MOEs) to excel in financial analysis tasks. By mimicking a real-world team setting, each model in FinTeamExperts specializes in one of three critical roles: Macro Analysts, Portfolio Managers, and Quantitative Analysts. This specialization allows the models to integrate their expertise effectively, forming a comprehensive and robust financial analysis tool.

Our approach includes several innovative contributions. Firstly, we are the first to implement role-based teams of LLMs as MOEs, aiming to mimic practical implementation scenarios within the finance domain. This method leverages the strengths of individual expert models while maintaining flexibility and adaptability in handling a wide range of financial tasks. Secondly, we introduced advancements in the MOEs architecture, such as dynamic routing, specialized training, and hierarchical expertise, which significantly enhance the model's performance in downstream financial tasks.

Through instruct-tuning and rigorous experiments, we demonstrated that FinTeamExperts outperform existing LLMs in finance-related tasks, underscoring the effectiveness of our pretraining methodology and specialized training strategies. These contributions showcase the potential of advanced LLMs in transforming financial analysis and decision-making, paving the way for more sophisticated and practical AI applications in the finance industry.

There are several directions for future exploration and enhancement. First, expanding the diversity of role-based teams within the MOEs framework could further refine task-specific expertise, particularly by incorporating specialized models trained on emerging financial topics, such as ESG (Environmental, Social, and Governance) criteria or digital asset analytics. Additionally, investigating the effects of cross-domain transfer learning may yield insights into how financial LLMs can benefit from knowledge in adjacent fields, such as legal or regulatory compliance, which often intersect with financial analysis. Another promising direction is the optimization of dynamic routing strategies to allow for more granular control over model selection based on real-time task complexity and data characteristics. This could involve developing adaptive routing algorithms that leverage reinforcement learning or other self-learning techniques, allowing the MOEs framework to continuously improve task assignment efficiency and performance.


\bibliographystyle{unsrt}  
\bibliography{references}

\end{document}